# Will bots take over the supply chain? Revisiting Agent-based supply chain automation

Liming Xu, Stephen Mak and Alexandra Brintrup*

Institute for Manufacturing, Department of Engineering, University of Cambridge

* Corresponding author, ab702@cam.ac.uk

**ABSTRACT**

Agent-based systems have the capability to fuse information from many distributed sources and create better plans faster. This feature makes agent-based systems naturally suitable to address the challenges in Supply Chain Management (SCM). Although agent-based supply chains systems have been proposed since the early 2000s; industrial uptake of them has been lagging. The reasons quoted include the immaturity of technology, a lack of interoperability with supply chain information systems, and a lack of trust in Artificial Intelligence (AI). In this paper, we revisit the agent-based supply chain and review the state of the art. We find that agent-based technology has matured, and other supporting technologies that are penetrating supply chains; are filling in gaps, leaving the concept applicable to a wider range of functions. For example, the ubiquity of IoT technology helps agents "sense" the state of affairs in a supply chain and opens up new possibilities for automation. Digital ledgers help securely transfer data between third parties, making agent-based information sharing possible, without the need to integrate Enterprise Resource Planning (ERP) systems. Learning functionality in agents enables agents to move beyond automation and towards autonomy. We note this convergence effect through conceptualising an agent-based supply chain framework, reviewing its components, and highlighting research challenges that need to be addressed in moving forward.

## 1. Introduction

Artificial intelligence (AI), also known as machine intelligence or computational intelligence, is used to describe machines that "mimic" and improve human cognitive abilities that humans associate with human intelligence, such as learning and problem solving (Russell and Norvig, 2009). The field of computer science often conceptualises AI as the study of "intelligent agents", which are computational programs that mimic and then act on behalf of a human user. Intelligent agents "sense" and "perceive" the environment in which they are located and take actions that can maximise the probability of achieving their "goals". They can use various tools to do so, including the ability to learn from data (Machine Learning), develop reasoning (Logic), communicate and negotiate with other agents (Multi-Agent Systems), evolve to adapt to changing circumstances and find solutions (search and optimisation). Each of these abilities have formed specialised AI sub-fields with different technical considerations.

AI and its impact on Supply Chains (SCs) has recently become a popular topic of discussion prone to hype, hope and fear (The Economist, 2018). Two central topics of interest surrounding the discussion of AI in SC are: data analytics and automation. Data analytics has been a key area of focus in many recent reviews, e.g., Wang et al. (2016) and Nguyen et al. (2018). While the boundaries between the two topics overlap in many applications, this paper mainly focuses on the latter. A recent study proposes that purchasing managers are at a 64% risk of losing their job due to automation, while buyers are at a 64% risk (Frey et al., 2016). In industries where purchasing costs can amount to over 50% of total costs, automation can become a key contributor to profit margins, making it an attractive option. In our context, the human whose behaviour is mimicked by an intelligent agent, is a SC professional.

In SC, the goal of a human-mimicking Intelligent agent would be to exhibit "smart" SC behaviour. The definition of what smart SC constitutes are various and goal posts move over time — however scholars generally agree that smart SC behaviour would involve better access and use of data for improved and automated decision making. With this in mind, we first reviewed the extant studies over the last 20 years on SC intelligent agents; showing the definitions of agent, agent-based system and simulation, and the distinction between agent-based system and agent-based simulation. We then reviewed the history of automation in supply chain management, and classified the reviewed studies in terms of automation level and supply chain functions. On the basis of the classifications of automation of agent technology being used in the SCM, we reviewed and discussed the convergence of agent technology and other new emergent technologies in SCM. To eventually adopt these mentioned technologies in SCM, system implementation and deployment and the factors that impact the industrial adoption are two concerning aspects for both academics and practitioners. We, therefore, reviewed the works within these two subfields and expected to reveal the obstacles that impede the industrial acceptance of agent-based technologies in the SCM.

The main purpose of this paper is to conduct a comprehensive review of automation by intelligent agent technology in supply chain management by analysing and consolidating studies that involve software agent methodologies to achieve supply chain goals. Our overarching research question is thus:

> *To what extent has research in the field of SCM exploited intelligent software agent capabilities and what are the opportunities and challenges that will likely define the future of the field next?*

While a small number of literature reviews on topics relevant to Agent-based Systems in Supply Chain Management (ABS-SCM) have been published, there has not been any attempt to review and categorise ABS-SCM literature to extract the field's current state, and define future research directions that need to be tackled in linking agents to supply chain automation. The main contributions of this work include:

- A detailed categorisation of ABS-SCM research areas spanning the fields of computer science and supply chain research;
- Identification of research gaps and relevant future directions for exploiting the current advancements in intelligent agent automation for SCM;
- A comprehensive analysis of ABS-SCM literature across supply chain functions, and levels of automation achieved;
- An investigation on the extent to which ABS-SCM has been integrated with digital supply chain tools such as IoT;
- An understanding of the relationship between ABS-related methodologies and performance metrics, and the SC tasks on which they are applied.

As succinctly presented in Olhager (2013), production research and especially planning and control, has witnessed several paradigm shifts since the 1960s, moving from a shop floor level to enterprise resource planning and eventually supply chain-wide decision-making. It can be argued that AI has gradually brought about a new paradigm shift, leading to automated systems that can harness knowledge and data to improve decision-making within supply chains. This survey contributes to understanding of the progression of this shift so far, in what concerns SCM, and mapping the way forward.

The rest of this paper is organised as follows. Section 2 gives a short background on intelligent agents and how they differ from simulation in helping drive real life information systems. Section 3 details the review methodology. Section 4 presents a broad classification of the reviewed studies along with bibliometric analysis. Section 5 synthesises the history of automation in Supply Chains. Section 6 reviews automation in terms of SC function. We then present the main research areas in agent-based supply chain automation, and industrial adoption, followed by discussion and conclusions.

## 2. Intelligent Agents and Agent Based Systems

We begin with a definition of intelligent software agents and related notions. Intelligent agents are autonomous computational entities that act in pursuing their own goals, with tools available to them. They react to the changes in their environment, reason, and can communicate with other agents and/or with human operators (Wooldridge and Jennings, 1995). The concept of agents and of MASs emerged from a number of research disciplines including Artificial Intelligence, Systems Design and Analysis using object-oriented methodology and human interfaces (Jennings, Sycara and Wooldridge, 1998).

Intelligent agents have a rich history in SCM, where they have been primarily used for simulation and modelling of SC dynamics, and for automating decisions and actions. Intelligent agent simulation has been used to model bottom-up supply chain emergence, where agents that represent suppliers are given goals such as minimisation of cost and inventory, and interact with other agents such as suppliers and buyers to achieve these goals. Some of the notable work include supply network configuration (Akanle and Zhang, 2008), risk management (Giannakis and Louis, 2011; Nair and Vidal, 2011), and supply chain collaboration (Fu et al., 2000).

Agent-based system (ABS) approaches in supply chains differ from agent-based modelling (ABM) and simulation. ABS involves the design and deployment of intelligent software agents that automate tasks using several AI paradigms, including prediction, decision making and actuating, and learning from decisions. Software agents represent a software development paradigm which is appropriate for distributed problem solving. The term "agent" here denotes an encapsulated software-based computer system that has autonomy, social ability, reactivity, and pro-activity and can cause changes in real life as well as interact with other software agents (multi-agent systems). A multi-agent system consists of a number of agents, which interact with one another in order to carry out tasks through cooperation, coordination and negotiation (Wooldridge, Jennings and Kinny, 2000).

In contrast, agent-based modelling (ABM), is a simulation paradigm where software agents are used to mimic supply chain behaviour for what-if analysis. ABM is often a prerequisite to designing ABS itself (see e.g., Oliveira, Lima and Montevechi (2016) for a review).

Being a real-life tool and not a simulator, ABS has distinct considerations that are critical for its safe and successful adoption. These have formed several subfields in Computer Science, some of which have been investigated in supply chain applications, such as agent architecture, negotiation and communication protocols, agent learning, and system integration. We will review these in the following sections.

Table 1. Comparison with related surveys. *Survey*: References; *Pub. year*: publication year; *Time period*: Time period the survey was conducted; *Shared studies*: How many papers reviewed are also being reviewed in this paper? *Focus area*: What as the focus area of the survey? E.g., agent-based models in the supply chain.

| Survey | Pub. year | Reviewed articles | Time period | Shared studies | Focus areas | Bibliometric analysis | Citation network analysis |
|---|---|---|---|---|---|---|---|
| Moyaux, Chaib-Draa and D'Amours (2006) | 2006 | 21 | 1993 - 2004 | 4 | Justify the use of agent technology in SCM; Review and compare 10 exemplified projects that use agent technology in SCM; | No | No |
| Lee and Kim (2008) | 2008 | 42 | 1996 - 2005 | 15 | Review the development and use of multi-agent modelling techniques and simulations in the context of manufacturing systems and SCM; | No | No |
| Dominguez and Cannella (2020) | 2020 | 66 | 1998 - 2019 | 7 | Briefly reviews the applications of MAS and agent-based simulation in SCM, including the software (MAS development frameworks) used for creating MAS, and the degree of development of the reviewed work. | No | No |

## 3. Review Methodology

### 3.1. Search strategy

The literature search started by defining the keywords used for collecting most relevant literature to our topics. Two levels of keywords are used combined with AND. The first level defines the search context and includes the term 'supply chain' OR 'supply network' OR 'logistics' OR 'procurement'. The second level includes the following terms related to intelligent software agents, as detailed in Section 2: 'intelligent agent OR software agent OR multi-agent'.

Google Scholar, which directly accesses journal and conference websites as well as patents, was first used. This was followed by a secondary search in Elsevier's Scopus and Clarivate Analytics' Web of Science. The earliest reviewed study was published in 1993. Hence, it was reasonable to set the time period from 1993 up to and including December 2020. This review methodology is illustrated in Fig. 1.

### 3.2. Search scope

A set of inclusion and exclusion criteria were selected to define the scope of the literature survey. These include:

- Studies must be peer-reviewed and written in English.

- Each study must include at least one practice, technique or methodology in the field of intelligent software technology. For example, supply chain management publications that include the word intelligent agent but have no concrete link to computational software agents are excluded.
- Studies are also excluded if the proposed approaches do not satisfy the ABS definition given in Section 2.1 and primarily focus on agent-based modelling and simulation.

The literature search yielded more than 200 results and the subsequent applications of the aforementioned exclusion and inclusion criteria resulted in 132 papers reviewed in this survey. This process of searching, and reviewing has been repeated several times between August and December 2020.

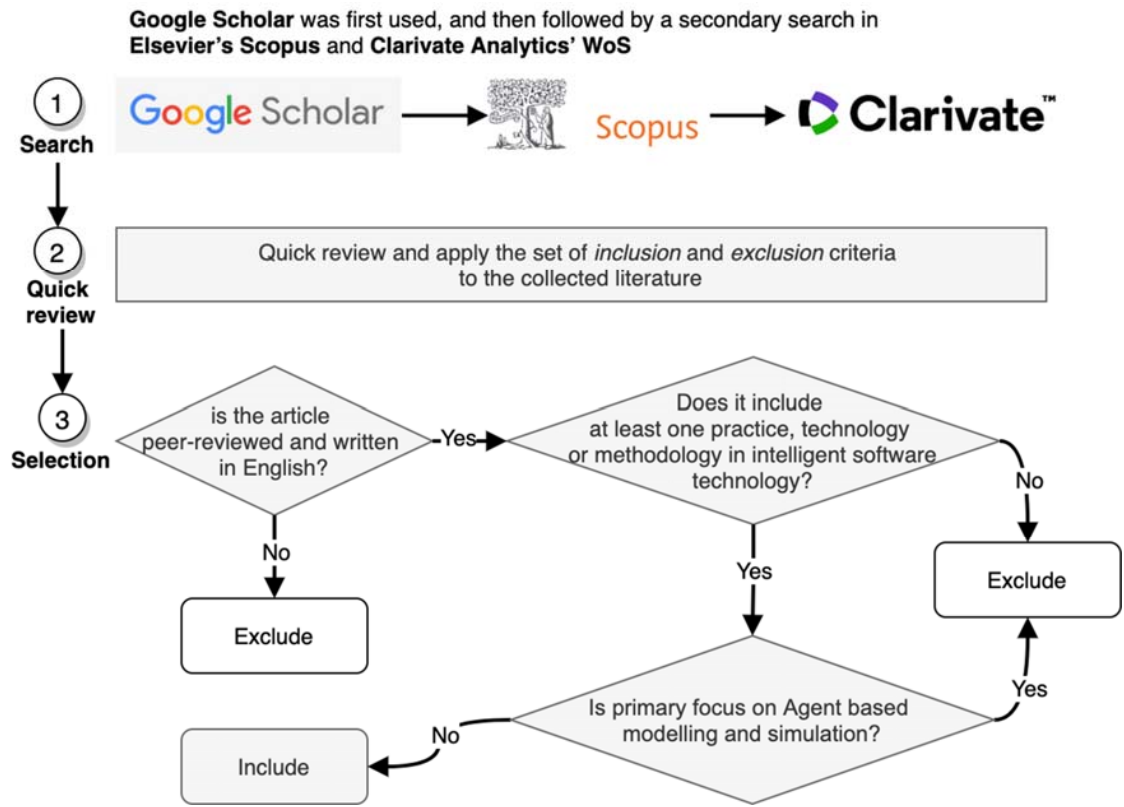

**Figure** 1. Depiction of the review methodology adopted in this article.

### 3.3. Related surveys

Despite the relatively short history of ABS-SCM as a distinct research field, a number of articles have been published with the aim of consolidating extant literature. In Table 1, we summarise the survey papers published in the past 15 years. Only those surveys that present a structured, systematic review process were included. The last comprehensive review took place in 2008, however the authors included multi-agent manufacturing systems as well as supply chain applications, whereas this survey focuses on the latter and provides an updated account since then, as well as providing specific discussion and guidelines for future research to advance the field. Additionally, we present and discuss operational workings of reviewed agent architectures and protocols. This paper further differs from extant surveys in that we exclude simulation only studies and focus on real life applications.

## 4. Classification of reviewed studies

### 4.1. Year of publication

The trend of publishing agent-related research in SCM shows a linear growth, starting with the first study in 1993 before a rapid increase to 10 publications in 2008, going back to 10 studies a decade later. There are noticeable sharp drops in 2000, 2007, and then a downward trend between then and 2017 before picking back up. Although not an exponential growth pattern, an increase in recent three years is encouraging and possibly motivated by the recent AI resurgence and interest in AI-focused SCM.

### 4.2. Venue of publication

Table 2 shows the top 5 publications venues ranked based on their contribution in the studies selected in this survey. Most contributions come from production, logistics and operational research journals, followed by journals that specialise in the intersection between Computer Science and production (e.g., Computers in Industry, Engineering Applications of Artificial Intelligence), and Computer Science outlets such as the AAMAS conference.

**Table 2**. Statistics for the reviewed literature with regard to publication type and venue

| Type of publication | Number | Publication venue | Number |
| --- | --- | --- | --- |
| Journal | 101 | International Journal of Production Research | 12 |
| Conference | 14 | Expert Systems with Applications | 9 |
| Workshop | 11 | International Journal of Production Economics | 7 |
| Book series | 4 | International Journal of Advanced Manufacturing Technology | 4 |
| Other | 2 | International Conference on Autonomous Agents and Multiagent Systems (AAMAS) | 4 |
| | | Others | 96 |

The field is notably dispersed in its publication outlets with no real, established repository that consolidates advancements. Some of the most highly influential articles in thinking have been computationally focussed, and published in outlets such as Association of Computing Machinery and

Applied Soft Computing. Whilst these have been referenced by subsequent computational research, they have not raised much attention in the operations research community, although this is the primary community the approaches are intended to be used by (for example, Singh, Salam and Iyer (2005) has a network citation of 0 but 75 in the larger computational community). One reason may be that traditional Operations Management (OM) journals are familiar with computer science terminology and their reviewer base does not involve relevant research expertise. This could of course be set to change as digital manufacturing technologies are becoming a significant differentiator of supply chain performance and computational research may penetrate further into operations.

*4.3 Bibliometric analysis*

We conducted a bibliometric analysis to extract major articles that influenced thinking in ABS-SCM. To do so, we investigated the number and patterns of citations. It is worth noting that the number of times an article has been cited is not directly related to the quality or the impact of the study, but it highlights the potential to influence readership of ABS-SCM, potentially leading further research directions.

Further bibliometric analysis included the creation of a citation network, which is a directed network where nodes refer to articles within our review, and links refer to citations referenced by the articles. Data was then analysed using Gephi 0.9.2 (Bastian, Heymann and Jacomy, 2009). It is thought that an increased number of shared citations between two publications points to the papers sharing a specific worldview (Boyack and Klavans, 2010). Hence, by performing a citation network analysis we aim to extract clusters of papers that share common threads of thought and build on each other. A highly disconnected cluster pattern would indicate dispersed areas of study that are not fully aware of each other.

As authors cite articles both within and outside our domain of inquiry (ABS-SCM), not all articles in the citation network are connected. To investigate citation behaviour within ABS-SCM, we extracted the largest connected component as shown in Fig 2.

Table 3 shows the top cited papers within the largest connected component and within the wider network. It is interesting that the two lists are not a perfect match. For example, the seminal paper of Maes, Guttman and Moukas (1999) "Agents that buy and sell" is the top cited paper in our review. The paper highlights the premise of creating automated auction and negotiation strategies through learning agents in e-Commerce. While this area is obviously related to the automation of procurement processes in SCM, it has been widely cited within the computer science community in studies that explore agent-agent negotiation, trust, and mobile agents. However, this paper has not been highly cited within the ABS-SCM network. Instead ABS-SCM researchers top cited paper is Swaminathan, Smith and Sadeh (1998)'s research on the modelling of supply chain dynamics with agent-based systems, which does not motivate the use of agents in real life transactions, but builds a modelling framework for simulation based studies, which then was used by subsequent SCM researchers to test how an automated agent-based supply chain would behave.

**Table 3.** Top cited papers (within the largest connected component and within the wider network) and their respective citations or scores.

| Top cited articles within citation network | Citations/ Score | Top bridging articles | Score | Top cited articles of all reviewed papers | Citations |
|---|---|---|---|---|---|
| Swaminathan, Smith and Sadeh (1998) | 1275 (22) | Lou et al. (2004) | 28.33 | Maes, Guttman and Moukas (1999) | 1275 |
| Fox, Barbuceanu and Teigen (2001) | 782 (18) | Jiao, You and Kumar (2006) | 22.50 | Swaminathan, Smith and Sadeh (1998) | 1267 |
| Sadeh et al. (2001) | 153 (12) | Akanle and Zhang (2008) | 14 | Fox, Barbuceanu and Teigen (2001) | 782 |
| Chen et al. (1999) | 182 (9) | Frayret et al. (2007) | 11.83 | Glushko, Tenenbaum and Meltzer (1999) | 502 |
| Fox, Chionglo and Barbuceanu (1993) | 254 (7) | Chen et al. (1999) | 9.00 | Giannakis and Louis (2011) | 348 |
| Yung and Yang (1999) | 54 (7) | Lo, Hong and Jeng (2008) | 8.50 | Julka, Srinivasan and Karimi (2002) | 303 |
| Kaihara (2003) | 220 (7) | Kim and Cho (2010) | 7.50 | Xue et al. (2005) | 286 |
| Lou et al.(Lou et al., 2004) | 113 (6) | Giannakis and Louis (2011) | 7.17 | Fox, Chionglo and Barbuceanu (1993) | 254 |
| Swaminathan, Smith and Sadeh (1996) | 69 (5) | Julka, Srinivasan and Karimi (2002) | 5.83 | Jiao, You and Kumar (2006) | 238 |

Cluster analysis shows 7 distinct clusters (obtained using the Blondel algorithm (Blondel et al., 2008) and highlighted with different colours in Fig. 2). Each article in the analysis is assigned to only one cluster. The largest cluster led by the work of Fox, Barbuceanu and Teigen (2001) focusses on negotiation and bidding strategies, and e-Supply. The second largest cluster has the largest number of studies in agent-based coordination, information sharing, and supply chain configuration, building on the work of Swaminathan, Smith and Sadeh (1996). Another emergent cluster contextualises work on forecasting, and inventory control. A distinct "application" cluster collates work on agile and sustainable supply chains, as well as specific cases such as forestry and fashion apparel. Finally, one small cluster building on the work of Bodendorf and Zimmermann (2005) builds on supply chain event management using agents to monitor events on social media and issue control mechanisms.

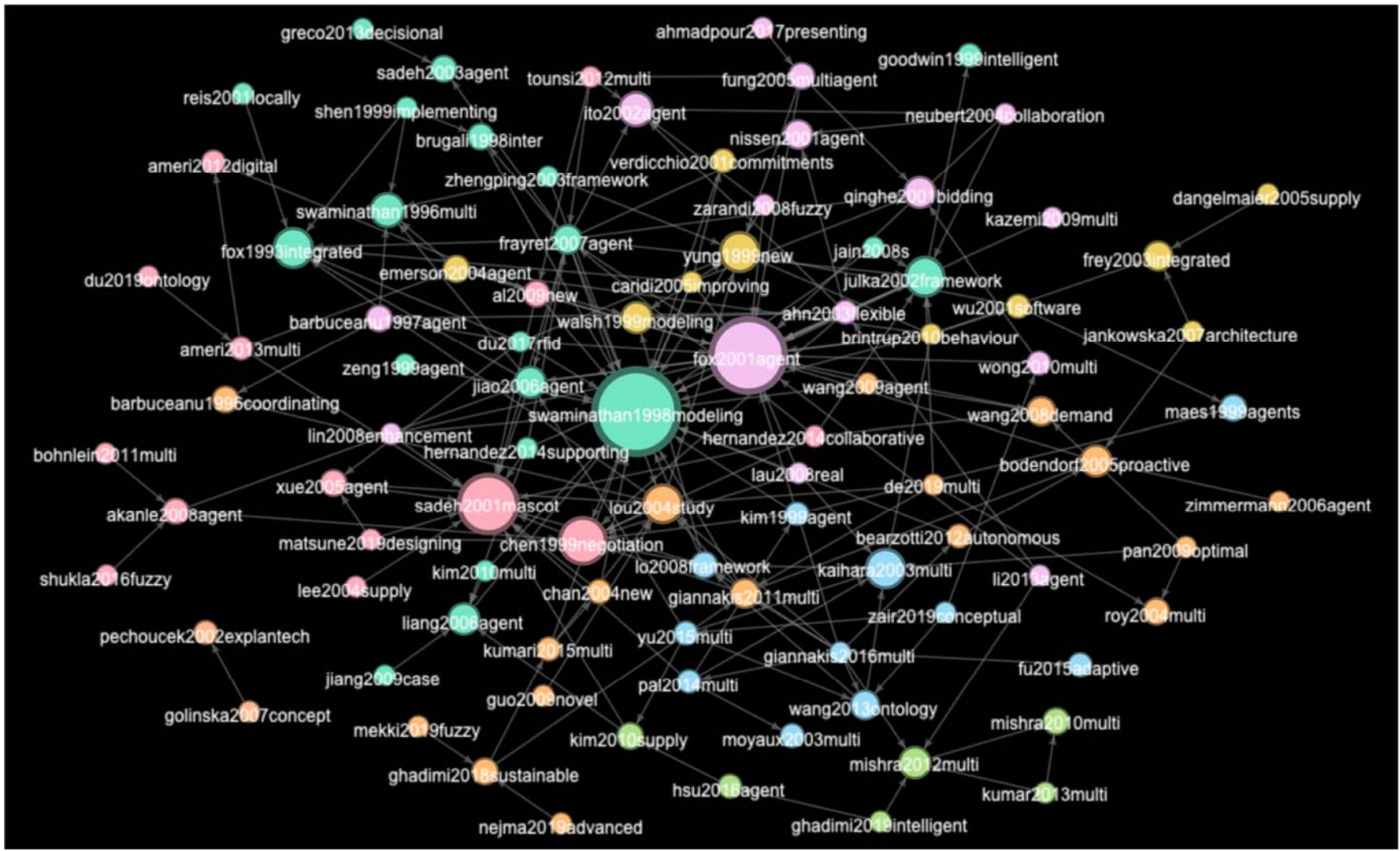

**Figure 2**. Citation network of ABS-SCM. Nodes are coloured according to clusters and sized according to their in-degree.

Looking at betweenness centrality, Lou et al. (2004)'s work on agile supply chains is the most central, acting as a bridge to connect different clusters in the network. Lou et al. (2004) defines a reconfigurable SCM system, by defining agent types and roles, which together share the task allocation using a contract net protocol. As this article develops a viewpoint that can be used for the focus areas of clusters on SC coordination, dynamic control and negotiation, it is cited by a variant of clusters. Similarly, the second article with high betweenness centrality, that of Jiao, You and Kumar (2006) also explores autonomous negotiation and collaborative planning and hence is cited by the two different clusters that independently investigate effective negotiation and coordination.

Top cited articles within the network are all from 20 years ago, whereas top bridging articles are more recent, where authors bring concepts from different focus areas together. Research in general is dispersed in clusters with a few key articles being centrally cited – which is also evidenced by the diversity of publication venues. The diversity of outlets and dispersion of focus areas within clusters may be one reason behind the slow progress of ABS-SCM and generally low citation numbers within the largest connected component.

## 5. A brief history of automation in SCM

In this section we consolidate research in ABS that resulted in the proposal of automated SCM functions. We first review the literature by the order of SC automation complexity levels, from low complexity to middle and then to high level automation complexity. We, then, generalise these three levels of automation complexity, assigning them with the corresponding terms: *process* automation, *operational* automation, and *tactical* automation.

The first steps of automation in SCs started in the 1990s with the now ubiquitous electronic procurement (e-Procurement) (Neef, 2001; Jonsson et al., 2011). Benefits that were highlighted included a reduction in purchasing costs, increased accuracy and speed of acquisition, reduced paperwork and administrative costs, and provision of better information for managers (Giunipero and Sawchuk, 2000; Deeter-Schmelz et al., 2001). Another benefit included potential expansion in customer base as more relationships can be easily handled without the need to further invest in human capital.

After the introduction of e-Procurement, the idea of automatically learning and devising strategies for negotiation on automated SC procurement platforms has gained interest. Given software agents' inherent suitability for these tasks, most follow-up works on automation in SCM used agent-based systems, which began with the proposal of several frameworks detailing agent roles and activities (Glushko, Tenenbaum and Meltzer, 1999; Julka, Srinivasan and Karimi, 2002; Xue et al., 2005; Jiao, You and Kumar, 2006). We will review these in the next section.

Agents have been used in Business-to-Business (B2B) environments for learning to bid effectively and arrive at trade-off solutions (Coehoorn and Jennings, 2004). Automated formation of ad hoc supply chains using agent-based systems (ABSs) were developed (Lou et al., 2004; Wang et al., 2009; Ameri and Patil, 2012; Ameri and McArthur, 2013; Greco et al., 2013; Shukla and Kiridena, 2016). Following these, automated inventory planning (Julka, Karimi and Srinivasan, 2002; Julka, Srinivasan and Karimi, 2002) and automated event monitoring (Bodendorf and Zimmermann, 2005) gained interest.

Further streams of work focussed agent use for SC coordination using ABS. These studies discussed various performance factors such as scalability, decision optimality, suitable for supply chains, proposing different agent models (Fox, Barbuceanu and Teigen, 2001; Reis, Mamede and O'Neill, 2001; Ito and Abadi, 2002; Wang et al., 2009). Wang et al. (2009) suggested that agent coordination should have different decision-making strategies at different planning levels. Agent communication and negotiation should be used at the strategic level for supply chain formation. While agents, at the tactical level, can employ argumentation to understand the preferences and constraints of each agent, agents at the operational level would adopt various strategies for preferences selection.

Automated formation of ad hoc supply chains using ABS were also widely considered (Lou et al., 2004; Wang et al., 2009; Ameri and Patil, 2012; Ameri and McArthur, 2013; Greco et al., 2013; Shukla and Kiridena, 2016). (Huhns, Stephens and Ivezic, 2002) designed a prototype software system that could automate the construction of SC and B2B processes. For a given B2B scenario, the prototype software system creates a Unified Modelling Language (UML) interaction diagram, extracts B2B conversations, which creates state machines for configuring agent behaviour in B2B transactions.

Brintrup et al. (2011) proposed a procurement system where aircraft parts were coupled with sensors and when degradation at the part level is predicted, agents that are connected to the sensors could automatically source replacements and schedule service by negotiating with supplier agents. Different agent architectural solutions were assessed in terms of scalability, optimal decision making, and system resilience.

The use of agents in automated inventory planning started with (Julka, Srinivasan and Karimi, 2002) which proposed a framework for agent-based SCM, and illustrated its application to a refinery supply chain (Julka, Karimi and Srinivasan, 2002). The primary objective of this framework was to unify data flows across the SC and create more optimal decisions on inventory management. Kumar et al. (2013) applied agent-based planning and coordination concepts to the wine supply chain and Du, Sugumaran and Gao (2017) applied the concept to prefabricated component supply chains. Hernández, Lyons, et al. (2014) introduced collaborative planning of inventory in multi-tier supply chains with the support of a multi-agent system. Lee and Sikora (2019) introduced the design, implementation and testing of an intelligent agent for handling procurement, customer sales, and scheduling of production in a stylised SC environment. The SC agent uses dynamic inventory control and various reinforcement learning techniques to adapt dynamically to the changing environment created by competing agents.

Another development is automation of SC event management. Bodendorf and Zimmermann (2005) initially proposed that ABS was especially suitable for proactively monitoring events related to SC disruptions. Bearzotti, Salomone and Chiotti (2012) developed a subsequent multi-agent supply chain event management system which can perform autonomous corrective control actions to minimise the effect of deviations in a production plan that is currently being executed. Kumari et al. (2015) proposed an automated self-adaptive multi-agent system, which can help SMEs to take appropriate actions to mitigate the uncertainty in SCs, and reduce idle time of machines with reduced human intervention. Blos et al. (2016) proposed agent-based risk management to mitigate the effects of disruptions autonomously.

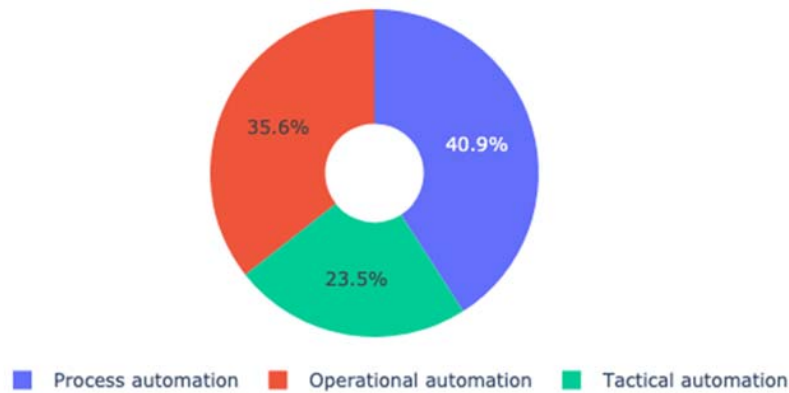

**Figure 3**. Automation in agent-based supply chain management

> Consolidating the above streams of studies gives rise to three distinct capabilities for automation in supply chains. The first one is the automation of low-level *processes* within supply chains such as data entry with Robotic Process Automation (RPA), and negotiation in eProcurement. The second level consists of automating processes in the *operational* time horizon, for example by coordinating production schedules and co-optimising inventory levels. The third level operates at the *tactical* horizon, for example by ad hoc supply chain formation and handling diverse B2B processes. Note that these capabilities are not defined by a "level" of intelligence, but rather the "complexity" of the process that the automation activity is being applied to. E-Procurement solutions may have sophisticated negotiation and pricing algorithms; coordination algorithms may be equally smart in utilising trade-off inventory optimisation solutions; and SC configuration algorithms may be intelligent enough to take into account multiple data and objectives from stakeholders. In all these levels, the intelligence technology used can be equally sophisticated, however, the time span of decisions ranges from a single procurement operation to the configuration of an entire supply chain. Most research reviewed above has focused on process and operational automation with tactical automation being the least explored area (see Fig. 3).

## 6. Automation across Supply Chain Functions

Following the standard Supply Chain Operations Reference model (SCOR), our analysis shows that the majority of ABS-SCM automation studies have been in the *Enable* functional area, followed by *Source, Plan, Make, Return* and *Deliver* (see Fig. 4). Enable includes processes associated with the management of: business performance, resources, facilities, contracts, supply chain network management, managing regulatory compliance and risk management. Most studies in this category used agent systems to automatically configure supply chains, and dynamically allocating orders to supply chain members. Planning functions that were automated included dynamic production planning as a result of orders from downstream supply chains (Lima, Sousa and Martins, 2006), and demand forecasts (Liang and Huang, 2006; Carbonneau, Laframboise and Vahidov, 2008), facility location (Kazemi, Zarandi and Husseini, 2009). Dangelmaier, Heidenreich and Pape (2005) investigated collaborative production planning.

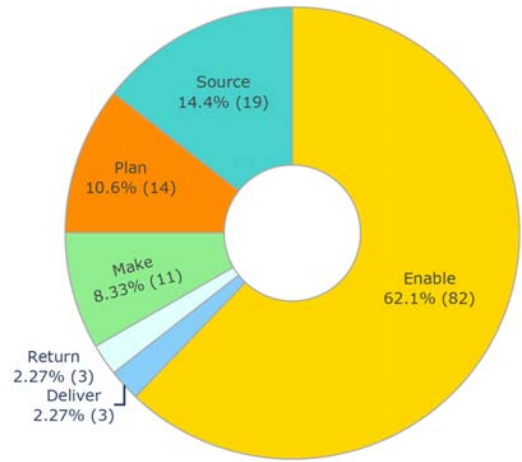

**Figure 4**. Distribution of supply chain automation articles within the SCOR model

The second largest category of articles have studied the use of agents in **Sourcing**. Sourcing studies mostly involved automated supplier selection such as Guneri et al. (2009), Yu and Wong (2015), Cavalcante et al. (2019) and Nejma et al. (2019). Zhang et al. (2015) created an agent-based peer-to-peer architecture for semantic discovery of manufacturing services across virtual enterprises, which was then used for supplier selection.

The third category involved **Planning** which involves SC processes that balance aggregate demand and supply to develop a best course of actions to meet sourcing, production, and delivery requirements. Here, Caridi, Cigolini and De Marco (2005) suggested an agent architecture for Collaborative Planning, Forecasting and Replenishment systems. Ito and Abadi (2002), Pan et al. (2009) and Kim and Cho (2010) similarly suggested ERP integrated agents to improve inventory control. Hernández, Mula, et al. (2014) proposed a decentralised multi-tier negotiation protocol which they piloted with an automotive case study. Here, independent MRP systems are able to negotiate through the use of software agents. Hernández, Mula, et al. (2014) showed that the automated negotiation led to better profit margins from both SC and individual organisation viewpoints. Kazemi, Zarandi and Husseini(Kazemi, Zarandi and Husseini, 2009) created a system that involved agents solving a distribution network problem using genetic algorithms.

Studies that fall under the **Make** category are those that integrate planning into production. Here researchers explored how supply chain scheduling can autonomously be linked to production scheduling (Wong et al., 2006; Aminzadegan, Tamannaei and Rasti-Barzoki, 2019), and how production orders can be automatically distributed among the supply chain (Reis, Mamede and O'Neill, 2001; Lima, Sousa and Martins, 2006; Gharaei and Jolai, 2018).

Only three studies have used multi-agent based automation approaches for **Delivery** and **Return**. Ying and Dayong (2005) proposed a framework for 3PL (Third Party Logistics) for the formation of ad hoc virtual private logistics teams, and built an automated route planning system for cases where transport and production schedules are closely connected and goods are distributed immediately after production. They applied this system to a German newspaper producer, reporting that their system was able to cope with fluctuations in production schedules better than the central planning system that was in place. Finally, Mishra, Kumar and Chan (2012), similar to Golinska et al. (2007), proposed a multi-agent architecture to handle recycling and reverse logistics issues, arguing that ABS provides an ideal solution to facilitate communication between disconnected parties in the reverse logistics chain. However, the system was not tested with real-world data.

> Mapping extant research onto standard Supply Chain SCOR functions reveals that the majority of historical research has used ABS for automation of *Enable* and *Source* functions, namely to select supply chain members, and assign orders. Automation of these functions are useful for fast, ad hoc supply chain formation, where longer term contracts may exist, but decisions at the operational time horizon are sped up. More recent work in this area has investigated automated agent search and semantic capability discovery. The rise of distributed manufacturing may popularise ABM - as manual ad hoc supply chain formation may not justify low volumes for custom products, hence encouraging ABS to be used for alleviating manual orchestration costs. Automation of supply chain formation naturally gives rise to distributed planning and negotiation, with direct input to ERP systems. We thus expect distributed planning to become a more popular area of scientific inquiry in SCM.

## 7. Main research areas in supply chain automation with agent-based systems

The main area of scientific inquiry that spans from multi-agent system based SC automation is the development of distributed AI solutions for relevant, specific SC problems, where autonomous information processing entities, i.e. agents, reach a consensus system state through interaction and communication. Hence the majority of research in SC-ABS investigated agent architecture, as well as the system architecture, which then involves distributed communication, negotiation and planning techniques (as shown in Fig. 5).

Research into agent architecture design is concerned with the distribution of tasks, and knowledge and control to multiple agents to solve a given problem. Once a given architecture is defined, the next level of inquiry concerns itself with effective coordination mechanisms, negotiation, conflict resolution, and communication protocols. The third area of research concerns with equipping agents with reasoning and problem-solving capabilities, with tools such as optimisation, and learning algorithms. Finally, the development of communication languages between agents and agent integration into enterprise software systems is needed to facilitate an environment in which agents can function and interact. In the following sections, we review these streams of research.

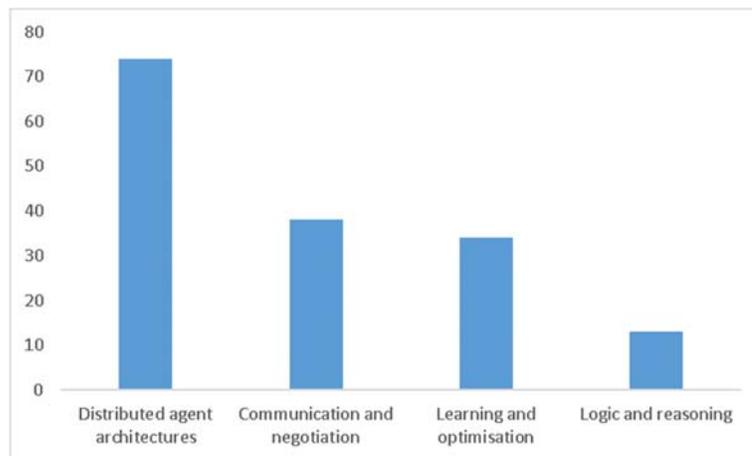

Figure 5. Main research areas in automation with ABS-SCM

### 7.1 Distributed agent architectures in ABS-SCM

Research into distributed agent architectures is concerned with the distribution of tasks, and knowledge and control to multiple agents to solve a given problem. Two general approaches for task distribution are followed:

*Functional decomposition* involves decomposing a supply chain task into functional agents that perform duties such as demand forecasting, production scheduling, logistics planning. Here, each agent takes responsibility into planning a single task and state variables are shared across agents. Depending on whether agents pursue their own goals or a system goal, they may negotiate or optimise system output towards a consensus state when conflicts arise across the functions.

*Physical decomposition* involves agents representing entities, which in the case of supply chains, may involve organisations, geographic locations. Here agents manage local state variables and similar to functional decomposition, agents negotiate or coordinate with other agents to carry out overall tasks.

Most architectures proposed for SCM have focussed on physical decomposition, where agents represent supplying and buying organisations, although some researchers that design ABS-SCM for a single organisation only have involved functional decomposition. Hybrids were also proposed where each organisational entity agent control over has sub-ordinate agents that functionally decompose tasks.

Following decomposition, the next stage is to define architectural options. In MAS architectural options are mainly categorised into *hierarchical, heterarchical* and *federated* (also called hybrid) architectures (Leitão and Karnouskos, 2015).

In hierarchical architectures information flows from lower levels to higher-levels until a suitable decision-making level is found, after which the decision is cascaded in the opposite direction (Leitão, 2009). The process is efficient for the collection and distribution of information from and to all levels, but creates bottlenecks when information needs to transfer between agents at the same level.

In a heterarchy, agents at the same level of the hierarchy can work together so as to react quickly instead of requesting control decisions from upper levels. In a fully heterarchical system there is only one-level where each agent can interact with another. Research has shown that such systems are robust and flexible to disturbances, however, reaching system-wide optimal decisions is not guaranteed, and system design needs to incorporate suitable cooperative mechanisms to avoid deadlock (Mařík and Lažanský, 2007). The move from hierarchies to heterarchies originated in traditional control systems (Trentesaux, 2009) with increased importance of flexibility and the need to represent agents that may have impartial knowledge of the system.

Federated architectures are designed to integrate aspects of hierarchical and heterarchical architectures into a distributed control system as a compromise for industrial agent-based applications. Here, different agents whose role it is to facilitate exchanges are introduced. These include: facilitators, brokers, matchmakers, and mediators (Weiming Shen, 2002). Facilitators convey messages between groups of agents by routing outgoing messages to the appropriate destinations and translating incoming messages to agents it is responsible for. Brokers are agents who find and procure specific services for agents, whom any agent can contact. Matchmakers, also called yellow page agents, may connect brokers to agents.

In ABS-SCM, while initial propositions involved heterarchy, most of the subsequent literature proposed federated architectures with physical decomposition at the enterprise level. Given the advantages provided by decentralisation, most ABS-SCM do not utilise hierarchical architectures.

Example architectures that combine different architectural arrangements and decompositions are as follows:

- Gjerdrum, Shah and Papageorgiou (2001) proposed a *hierarchical* arrangement with *physical* decomposition. Here a customer agent simultaneously interacts with warehousing agents, who in turn may pull orders from factory agents which then task transport agents with delivery.
- Lo, Hong and Jeng (2008) combine *physical* and *functional* decomposition with a *federated* architecture for a web-based procurement system for the fashion industry. Functional decomposition includes buyer and supplier agents while physical decomposition includes wrapper agents to CRM and ERP software.
- Wang, Wong and Wang (2013) propose a flexible *federated*, *physically* decomposed agent architecture for negotiation between supplier and buyer agents that rely on heterogeneous learning and knowledge representation schemes. They develop a negotiation ontology to facilitate communication. Agents find each other through a directory facilitator (matchmaker agent). If multiple entities exist on the buyer and seller side, their respective position is coordinated via a coordinator agent.
- Shukla and Kiridena (2016) propose a dynamic supply chain configuration framework to reduce transport carbon emissions and cost. However, the configuration is viewed through the lens of a single supply chain designing firm. They follow a *federated* architecture with *functional* decomposition, involving knowledge acquisition and representation agents, configuration predictor and evaluator agents, followed by a dispatch agent that actuates decisions.
- Giannakis and Louis (2016) propose a heterarchical agent supply chain framework that is *functionally* designed for "big" data processing. Each supplier has an order management, coordination, production planning, inventory management, contract management, logistics and monitoring agent as well as a "big" data agent. Inter-organisational communication is done via the supplier's communication agent. Any supplier can contact any other. The big data agent utilises case-based reasoning to suggest actions upon analysing internal and external data. The framework has not been implemented in real life and it is unclear why case-based reasoning is proposed for data analysis.

It appears that automated bidding problems in electronic procurement markets tend to focus on heterarchical communication where sellers and buyers representing their organisations (functional decomposition) communicate directly, with some form of federation involved for agents to find each other (e.g., with a yellow page agent).

In contrast, in more holistic supply chain planning and coordination problems, such as inventory planning across multiple tiers, functional and physical decomposition are combined where different agents are responsible for knowledge acquisition, learning and planning, after which coordination agents communicate decisions between organisations.

However, it is largely unclear why certain architectures have been preferred over others, what the refactoring process for a particular architecture has involved, and how architectural choices would perform for different supply chain problems. Few papers draw comparisons and test them with performance metrics. We will discuss these issues further in the following sections.

*7.2 Communication and negotiation protocols used in ABS-SCM*

After architectural design is defined, protocols for coordinating agent – agent communication must be introduced. The most common MAS protocols include the *contract net protocol*, and *automated negotiation*, which help agents communicate task allocation in supply chain processes. Lee and Kim (2008) listed which communication protocols and architectures were used by SCM researchers until 2008. Here we detail their operational insights as well as bringing insights from computer science, and add further work done since then.

The earliest methods for agent task allocation included simple message passing (e.g., Swaminathan, Smith and Sadeh, 1998) and Contract Net Protocol (CNP) developed by Smith (1980). CNP includes a call for bids by a manager (contractor) agent, collection of bids from contractees and the subsequent assignment of the tasks (i.e., their acceptance or rejection). In the CNP applications, a contractee agent cannot consider several tasks at the same time and if a better task comes along, it will break off its prior commitment, which results in the manager to recall an auction for the broken tasks.

Sandholm's seminal paper (Sandholm, 1993) extended the CNP and can be considered a first application into supply chain management, as the protocol has been implemented to the TRACONET (TRAnsportation COoperation NET) system, where dispatch centres of different companies cooperate automatically in vehicle routing. In his paper the CNP is modified such that agents are able to cluster announced tasks and can choose the stage and the level of commitment dynamically. If the agent chooses to break a contract, the manager penalises the agent's reward. Aknine (Aknine, 1998) further modified Sandholm's approach, and proposed that an agent can bid for several tasks at the same time without running the risk of being penalized in case he breaks the contracts. Here there are pre-Bidding, definitive-Bidding, pre-assignment and definitive-Assignment stages. They applied the extended protocol to scheduling goods delivery. Compared to the traditional one, the duration of the negotiation is much inferior as the number of messages exchanged increases, however more realistic solutions emerge. Interestingly, these extensions, despite being applied directly to SCM, have not been noticed by subsequent SCM researchers, but they have been popular in the field of computer science for other applications. SCM researchers that have used CNP include Petersen, Divitini and Matskin (2001), Ahn, Lee and Park (Ahn, Lee and Park, 2003), Kotak et al. (2003), Roy et al. (2004), Tah (2005), Ulieru and Cobzaru (2005) and Lu and Wang (2008). Lau et al. (2005) proposed a modified CNP for the distributed scheduling of supply chain deliveries. The motivation is that suppliers in conventional CNP cannot improve bids as there is only a single operation for bidding at a time. They relaxed this constraint but raised that a time window during which a bid is valid needs to be determined, so that agents would not wait indefinitely for receiving messages.

Lin, Kuo and Lin (2008) viewed the order fulfilment planning problem in the metal industry as a distributed constraints satisfaction problem and solved it with an asynchronous weak-commitment search protocol. Here, a solution that fulfils all agent constraints is sought through. If a local constraint is violated, for example an order cannot be fulfilled, the agent that is violating the constraint initiates a prioritised resolution process – for example by searching other agents that it can outsource the task to. Their solution can be viewed as a modification of the CNP. Akanle and Zhang (2008) proposed an iterative bidding mechanism to configure the supply chain which can then be used to group frequent formations into individual clusters. Lopes, Novais and Coelho (2009) created a software agent framework for automated supply chain negotiation, discussing the suitability of classical negotiation paradigms, and considering how design and coordination problems could be addressed.

Next SCM researchers tried to incorporate multiple criteria in the CNP. Wong and Fang (2010) created the ECNPro (the Extended Contract-Net-like multilateral Protocol), which is designed to handle multi-issue SCM negotiations in which the buyer is concurrently negotiating with many suppliers. They deploy a Multi-Attribute Utility Theory (MAUT) approach to establish the utility functions for a set of negotiation issues in the bargaining process. Their approach can split an order to more than one supplier to achieve a better negotiation payoff. Similarly, Kruse et al. (2013) used a double auction protocol in an agent driven service supply chain system where a multi-criteria function is used to convey overall value from multiple parameters.

Risk and uncertainty have also been a research theme in communication protocols. Bearzotti et al. (2012) extended CNP for SC event management, where a set of resources representing agents to interact through an order agent. Wang, Wong and Wang (2013) built a tactical bidding model of both parties, which take into account the impact of different risk preferences and discuss how different risk preferences affect the negotiation number, price, and utility of agents. Hernández et al. (2014) created a negotiation-based agent system for collaborative replenishment in a multi-tier supply chain that creates plans in the one-year planning horizon, where the negotiation process aims at reaching a consensus by adjusting plans within a maximum price limit for the end product. Manupati et al. (2016) proposed a mobile agent approach where agents travel to a different enterprise system to take on the task of negotiation for job scheduling. Hsu et al. (2016) presented an agent negotiation mechanism that combines fuzzy logic with negotiation strategies (e.g., competitive and win-win) for SC planning (see also Morganti et al 2009).

In addition to negotiation protocols, work has been done to learn from negotiations and form negotiation tactics. In this stream of research, Kraus, Wilkenfeld and Zlotkin (1995) and Kraus (2001) first proposed a generic multi-agent negotiation model, which explores the multi-agent negotiation under incomplete information conditions.

However, this earlier work in learning from agent negotiation took a while to transfer to SCM. Coehoorn and Jennings (2004) were amongst the first to propose a system to learn the opponent's preferences to make negotiation trade-offs in e-Procurement. Their proposal used a generic Kernel Density Estimation based learning embedded in intelligent agents representing suppliers. Several follow-up works applied learning algorithms such as Q-learning and Bayesian methods to learn from opponents' bidding patterns. Sundarraj and Mok (2011) considered human elements (namely, goal, situational power, and learning) in automated SC negotiation, and proposed methods to model goal and situational power and showed how they can integrate with algorithms for learning a counterpart's negotiation tactics. Hernández et al. (2014) and Chen and Weiss (2013) developed automated negotiation strategies that can adjust agent utility by acquiring an opponent model through game theoretic arguments. Chen and Weiss (2015) proposed a bilateral multi-issue negotiation approach that enables an agent to effectively model opponents in real-time through discrete wavelet transformation and nonlinear regression with Gaussian processes. Wang, Wong and Wang (2013) proposed an ontology-based approach to organise the negotiation knowledge that is utilised by agents that can be learned.

While the field of computer science has a rich history of agent negotiation research, the field of SCM has not fully utilized these approaches. Further, similar to our findings in literature that designs SCM agent architecture, we have found that there are no research attempts to map which negotiation protocol could be suitable for different supply chain applications. Further, research that designs communication protocols for agents and research that designs agent architectures remain separate from each other. These two strands of research need to inform one another as in a real-life supply chain setting confidentiality issues become important, as does the scalability of message exchange with an increased number of suppliers taking up agent-based solutions.

*7.3 Learning and Optimisation*

The use of optimisation has been the third most widely researched area in ABS-SCM. Here, individual agents are given optimisation capability, typically a heuristic, to find the best course of action to take with the information available to them at the time. A key feature of agent technology is its ability to encapsulate a variety of other AI tools, with agents learning to solve problems, maximising individual or system level goals and exploring trade-offs in between. Agents can also use prediction to make estimations about the future and use this in decision making. This encapsulation is particularly appealing as the use of AI technology in supply chains has seen promising advantages in recent years in a range of problems from demand forecasting to routing optimisation.

The distinction between learning and optimisation presents a thin line in our context. It is important to note that learning algorithms embedded into agents themselves often use optimisation to refine algorithmic parameters, such as weights on a neural network, or in planning algorithms where a tree of goals and subgoals are searched to find a feasible path to an end goal. Similarly, reasoning tasks can be performed with search and optimisation, where an inference rule is applied iteratively through a tree of rules to search for a logical answer to a question.

While mathematical optimisation methods are not included under the taxonomy of AI, metaheuristic, nature inspired, and computational intelligence-based algorithms such as hill climbing, simulated annealing, tabu search, genetic and evolutionary algorithms, firefly, ant colony, artificial bee colony, and swarm intelligence algorithms are all considered to be types of AI based optimisation algorithms. Within this broad heading of optimisation used by agents, the vast majority of ABS-SCM publications used genetic algorithms, followed by other heuristic methods.

Supply chains are typically characterised by individual goal seeking agents, who must coordinate together to arrive at globally feasible solutions. In ABS-SCM, supply chain agents can work towards optimising individual goals, or collectively optimise towards a better system state.

For example, individual objectives could include inventory cost minimisation whilst a global objective might be lead time minimisation. The choice of selfish and collective objectives and their weighting would determine the quantity and location of inventory that should be produced, stored, and schedule with which material should flow between locations. When supply agents seek individual rewards without paying attention to the collective system goals, individual optimisation and learning algorithms are used. When a common objective is pursued, then collective intelligence would emerge.

Researchers have experimented with a variety of collective and individual goal seeking behaviour in ABS-SCM. Several authors sought to automate planning by combining both individual and global agent objectives. For example, Fung and Chen (2005) proposed a global objective function to minimize the sum of purchasing, transportation, inventory holding and penalty costs, while satisfying orders. Similarly, Pan et al. (2009) used a central inventory optimising agent in the garment supply chain that trades off individual agent goals. Chen and Cao (2020) used a global optimisation approach to combine multi-agent objectives in the recycling network planning problem, constituting a six-level closed-loop supply chain network. Gharaei and Jolai (2018) investigated the trade-off between reduction of cost with batch production and the increase in tardiness when items in the batch need to be delivered to multiple downstream customers. Here customers are represented by agents with competing goals, and a Pareto optimal solution is offered to obtain a trade-off. de Souza

Henriques (2019) used a combination of individual and global objective functions. In their system, each supplier and manufacturer maximises its own profit, while a global satisfaction objective ensures that delays are minimised. They conclude that in competitive environments the system will not be balanced.

Others proposed individual objective optimisation, with data sharing to encourage replanning so as to mitigate consequential effects on other agents such as the bullwhip effect (e.g., Yung and Yang (1999) and Zarandi, Pourakbar and Turksen (2008)). In situations where supplier goals do not conflict with one another, individual objective optimisation may yield globally optimal solutions. For example, Lim, Tan and Leung (2013) proposed a digital currency based iterative agent bidding algorithm to negotiate a multi-site production and delivery scheduling. A genetic algorithm is employed to tune the currency values for agent bidding. Here, each agent has individual objectives and a global objective. The global objective is to obtain a process plan and schedule that gives the lowest production cost while satisfying all requirements such as due dates, while the machine agent's objective is to optimise its machine utilisation. In the work by Lim, Tan and Leung (2013), agent goals are thus not contradicting one another and thus a global solution is possible.

The coupling of learning methods into agents for predicting the best course of action given a historical context is a less well researched area in supply chains, although some recent publications have emerged. Historically, learning has been used in negotiation, with several authors proposing to learn from opponents' bidding patterns (Coehoorn and Jennings, 2004; Jiang and Sheng, 2009; Chen and Weiss, 2015) but these approaches seem to have been siloed in the field of computer science. Jiang and Sheng (2009) and Kim, Kwon and Kwak (2010) used reinforcement learning for dynamic inventory control. Both research teams took a distributed problem-solving perspective, where the reward function is common across the agents, but each agent has its own span of control. More recently, (Kosasih and Brintrup, 2020) experimented with reinforcement learning agents to optimise safety stock, finding that agents collectively arrived at more innovative solutions than prescribed stock policies in literature. Predictive capability in agents have been seldom researched in ABS-SCM although predictive agents form a large field of inquiry in computer science. We envisage an increase in both learning and predictive agent capability given the increasing strengths and use of machine learning algorithms.

*7.4 Logic and Reasoning*

The next area researched in ABS-SCM is agents' use of logic and reasoning. Here agents use Logic and Reasoning for knowledge representation and problem solving. Automated reasoning techniques rely on a priori encoding of expert knowledge in the form of rules or cases. For example, Fuzzy Logic (FL) is a many-valued logic used in cases where truth values can be vague, rather than binary. FL has been widely used in coding expert judgement into agents, for example in supplier selection (Ghadimi, Toosi and Heavey, 2018) and supply chain configuration (Shukla and Kiridena, 2016). Qualitative supplier criteria such as "relationship closeness" or "reputation" can be coded as linguistic variables so that they can be added into an optimisation problem and be considered alongside quantitative variables such as cost and constraints such as capacity. Constraints themselves can be fuzzified for negotiation-based decision problems such as multi-supplier scheduling (Hsu et al., 2016). Lou et al. (2004) and Pal and Karakostas (2014) used case based reasoning methods for creating agile response strategies in the supply chain and for enabling service matching between customers and suppliers. However, researchers raised that FL and Expert systems do not scale well, and its opponents advocate the use of computational intelligence techniques that do not depend on a priori symbolic encoding.

*7.5 Performance metrics*

There exist no commonly used performance metrics for comparing architectural choices, negotiation protocols, and optimisation and reasoning tools suggested for ABS-SCM. Some papers evaluate proposals with metrics whilst many others do not. Metrics that have thus far been used include:

- Number of messages exchanged as agent numbers are increased (Brintrup et al., 2011);
- Optimal solutions reached given a certain timeframe as agent numbers are increased (Brintrup et al., 2011);
- Improvement in solutions compared to a benchmark (Shukla and Kiridena, 2016);
- Individual and aggregated utility at the end of negotiation round (Wang, Wong and Wang, 2013);
- Collaborative efficiency assessed via experts (Fu and Fu, 2015);
- Time taken to reach consensus (Bearzotti, Salomone and Chiotti, 2012);
- Time taken to reach consensus as agent numbers increased (Brintrup et al., 2011).

Further, no benchmark problems are reported in ABS-SCM literature. It is thus hard to evaluate how a given architectural paradigm, communication protocol or optimisation approach presents an advance to the field. In order to advance ABS-SCM, benchmark problems and common performance evaluation metrics appropriate to automated SC tasks need to be defined.

> This section reviewed primary research areas within ABS-SCM. Whilst the overarching research inquiry within ABS-SCM has focussed on the distribution of tasks, knowledge and control to multiple interacting agents within a given SCM problem; researchers have used a variety of approaches differing in their granularity. First of these has been architectural organisation, which defines agent types, and tasks within a system. Here physical decomposition has been the most popular, where agents represent entities and geographic locations, although hybrid approaches that combine functional and physical decomposition are emerging due to scalability concerns. Similarly, whilst initial propositions involved heterarchy, most of the subsequent literature proposed federated architectures with physical decomposition at the enterprise level. This is particularly the case when holistic use cases are concerned that are planning focussed.
>
> After architectural designs are designed, communication protocols are introduced. The most commonly used MAS protocols included the now-classical contract net protocol and automated negotiation, which help agents converge on task allocation in supply chain processes. Interestingly, various more realistic extensions to these have been proposed in the field of Computer Science, but these have been largely ignored in the ABS-SCM literature, despite having been proposed for enterprise use cases. These include multiple bidding, multi-criteria CNP, mobile agent integration to ERP, and learning from negotiation.
>
> The issue of collective versus individual goal-seeking is an important area of study that is both very much at the heart of many SCM problems and is at the same time a design decision in ABS. Here researchers have experimented with both common goal seeking, selfish behaviour as well as attempts for mediation through constraint based methods, and voting.  As learning systems such as Reinforcement learning become popular in ABS, more research needs to be carried out to understand how such schemes can be carried forward, and what the implications on SCM use cases would be.
>
> Finally, the field of ABS-SCM suffers from a lack of common performance metrics, benchmark datasets and open software, which causes research to stall.  Researchers are urged to make their efforts publicly available.

## 8. Software engineering for facilitation of ABS-SCM

This section considers extant ABS-SCM work from a practical lens, including software engineering for facilitating ABS-SCM, communication languages, and integration within existing supply chain information systems.

### *8.1 Communication languages*

Key to the agent-to-agent interaction is the use of appropriate, formal communication language. The most commonly used communication language between agents is the Knowledge Query and Manipulation Language (KQML) developed by Finin et al. (1994), which led to the high-level Agent Communication Language (ACL) proposed by Foundation for Intelligent Physical Agents (FIPA) as a standard.  KQML is an auction-based market protocol, used in ABS-SCM for supply chain formation and reconfiguration (Barbuceanu and Fox, 1995; Ahn, Lee and Park, 2003; Lou et al., 2004). (Ahn, Lee and Park, 2003) raised the limitation of existing agent conversation systems is that it is difficult to make agent-based supply chains adapt to new products or new trading partners because existing systems usually use a fixed set of transaction sequences (Ahn et al., 2002; Freire and Botelho, 2002). Viewing this as a critical barrier to adoption, they suggest a flexible conversation model which consists of an interpretable and exchangeable conversation policy model such that agent systems are enabled to acquire new conversation patterns from counterpart agents when changes occur.

Singh, Salam and Iyer (2005) proposed the incorporation of Web Ontology Language (OWL) basis to build more AI inspired knowledge representation languages that are unambiguously computer-interpretable, making them amenable to agent interoperability and automatic reasoning techniques. They suggest that accepted business ontologies can provide discovery and transaction facilitation for participants and help create intelligent e-marketplaces.

Further, Multi-Agent Logic Language for Encoding Teamwork (MALLET) was developed by Xiaocong Fan et al. (2006) to encourage team-oriented programming. Based on the information needed, this agent language framework facilitates and manages agent activities via the proactive exchange of information. MALLET facilitates the encoding of knowledge (i.e., declarative and procedural) and information flow in the system.

### *8.2 Integration with Digital Supply Chain Information systems*

Integration into Supply Chain Information systems is key to the success of seamless ABS-SCM as well as presenting a significant opportunity to facilitate the integration of inter-supply chain heterogeneous software and hardware systems, potentially leading to plug-and-play systems. Several past papers on e-commerce seem to have been developed in proprietary environments although few authors tested ABS implementations that integrate with SC enterprise resource planning (ERP) applications.

Turowski (2002) suggests integration of SMC-ABS with an ERP to XML data conversion agent. In his system when demand signal is automatically signalled to suppliers, ERP generated demand is converted into XML and transferred to the supplier using standard Internet protocols (TCP/IP), where the supplier agent performs another conversion into its own ERP system using an agent. Then, the request for a quote is processed by the supplier's ERP system resulting in an automatically generated bid, which is transferred to the manufacturer using the same mechanism.  However, in this system, changes to the individual ERP systems may result in disconnection so the system is not easily adaptable.

Chan and Chan (2004) argued that an agent framework can be treated as a wrapper application, and an interface can be added in the layer between the existing system and the agent framework. This method circumvents the requirement that supply chain members need to develop a new local information system (e.g., ERP). If this interface is designed as a web application, various parties can more easily coordinate distributed information.

Similarly, in 2007, in a joint research project Mobile Internet Business[1] funded by the German Federal Ministry of Education and Research, Jankowska, Kurbel and Schreber (2007) showed how heterogeneous systems can be integrated via web services, implemented in Compiere, an open-source ERP system. In their proposal, wrapper agents running within the JADE-based MAS provide access to the data layer. The business logic layer of the MAS is represented by resource, monitoring, user, and gatekeeper agents. Gateway agents represent the web services layer. While this system is inherently more flexible, they raise trust/security issues.

Despite these attempts, Giannakis and Louis (2016) highlighted that the development of solutions for holistic cross-organisational collaboration progress slowly. Significant amounts of time and funds have to be invested to transform conventional e-business systems into collaborative SCM systems, and therefore only large organizations can afford the investment to achieve responsiveness and strengthen their competitive advantages. Small parties in a supply chain would not gain benefits from e-business due to this constraint, even if ERP systems can handle extended enterprises. Giannakis and Louis (2016) suggested ABS as an alternative means of communication.

Emergence of a number of new digital technologies present an opportunity for more seamless integration of agent-based systems in supply chains, and a number of early works have discussed these. Internet of Things (IoT) technology allows a physical object to be bound to digital information about the object using unique object-level identification. A typical IoT architecture would be composed of four layers: sensors and identification, network, service, and interface layers. Research has investigated how software agents can access sensor data through IoT (Brintrup et al., 2011). This is relevant when the condition of a product that travels along supply chain locations needs to be monitored. Another area of growing relevance is Digital Twin, which is a collection of digital data representing a physical object. By bridging the virtual and the physical world, data is transmitted seamlessly to allow the virtual entity to exist simultaneously with its physical counterpart (Grieves and Vickers, 2017). Since the concept and model of the digital twin were publicly introduced in 2002 by Grieves (2016), it has been widely acknowledged in both industry and academia including manufacturing and supply chain management. The convergence of agent-based systems and digital twins in SCM has been explored in literature, referring to the recent work by Orozco-Romero et al. (2020) for more information.

Additionally, as the price of Bitcoin (Nakamoto and Bitcoin, 2008) continually reached new highs, blockchain technology has been increasingly gaining attention. Distributed ledger technology (DLT), an ancestral form from which blockchain (or particularly Bitcoin) evolved, has a rich past. The idea of distributed ledger traces back to the work by Haber and Stornetta (1991), which proposed practical procedures for certifying when a digital document is created or modified. Then the follow-up works such as Mazières and Shasha (2002) and Szabo (2008) facilitate the birth of Bitcoin and later on the other forms of DLTs (e.g., Ethereum (Wood, 2014), R3 Corda (Hearn, 2016) and Fetch Ledger (Baykaner and Rønnow, 2019)). The distributed nature of both DLT and agent technology would promote the intersection of these two technologies. Hybrid decentralisation via agent based blockchain solutions are being investigated as part of important initiatives such as the recent

---

[1] http://mib.uni-ffo.de

European GAIA-X data infrastructure initiative, which in turn is being trialled in use cases such as the automotive sector. We anticipate an emergent convergence of DLT and agent technology and subsequent trials within SCM.

> Whilst agent-agent communication protocols have been standardised, language research in SCM has not been a popular topic of inquiry, despite being raised as a limitation. We view this as a critical barrier to adoption, as many SCM problems will necessitate flexible conversation models that can adopt over time. Whilst a number of ERP-agent integration techniques have been proposed, there are no existing comparative benchmarks to improve upon. A number of ERP software such as SAP offer agent integration tend to be bespoke and not readily available. The current economic drive towards decentralisation evidenced by recent interest in blockchain and agent integration, may encourage further attention to overcome communication and system integration barriers in ABS-SCM.

## 9. Industrial adoption

A diverse range of industries have taken up interest in agent-based system use. These have included high value manufacturing such as automotive, aerospace and medical devices (Brintrup et al., 2011; Hernández, Mula, et al., 2014; Ghadimi et al., 2019), textile and fashion industries (Brugali, Menga and Galarraga, 1998; Lo, Hong and Jeng, 2008), energy markets (Lopes, Novais and Coelho, 2009), electronic devices (Ahn, Lee and Park, 2003), and chemical processing (Julka, Srinivasan and Karimi, 2002). Retail focussed industries have primarily focussed on exploiting agent systems for supply chain formation, data integration and associated inventory planning (Ito and Abadi, 2002; Piplani and Fu, 2005; Kim, Kwon and Kwak, 2010); whereas high value manufacturing industries concentrated on collaborative production planning (Dangelmaier, Heidenreich and Pape, 2005), with servitised manufacturers exploiting agents for maintenance supply chain planning (Zhang et al., 2015). Rapid, automated negotiation has been a factor for studying agents in energy supply chains (Lopes, Novais and Coelho, 2009).

However, despite its rich history and several benefits, ABS technology had not been widely adopted by industry by 2005 (Mařík and McFarlane, 2005). Mařík and McFarlane (2005) gave several reasons: cost of deployment being higher than centralised solutions; a cultural preference in engineering towards centralised solutions; a lack of agent standards for knowledge exchange and representation between virtual organisations; scalability; a lack of commercial solutions; trust and security issues. A follow-on study by Leitão and Karnouskos (2015) a decade later investigated a set of factors that impact the acceptance of industrial agents, including design, technology, intelligence/algorithms, hardware, cost, standardization, application and challenges, and concluded that a decision on agent utilisation in productive industrial systems is a complex undertaking and many issues still need to be addressed for achieving a wider industrial acceptance. However, as discussed in the work by Leitão and Karnouskos (2015), the emergence of industrial internet technologies, IoT, cloud systems and distributed intelligence may provide an opportunity to the adoption of agents in industrial sectors.

As most SC information systems are proprietary, it is difficult to say which ERP functions are being automated by agent technology, and surveys with SC managers are not helpful as low-level software functionality is not transparent to them. However, it would be reasonable to say there is awareness and ongoing pilots in large ERP system providers and that some of these are being taken up by companies. For example, SAP has demonstrated a prototype agent capability in its replenishment software. Here agents predict the probability of stock-outs based on current inventory, scheduled receipts and expected demand—and when that probability exceeds a certain threshold, a replenishment signal is triggered. Agents, hence, are reactive to predictions on inventory levels. P&G has then deployed agents in its SC planning through its SAP implementation. Oracle is another system provider that has incorporated agents into its Cloud ERP solutions, highlighting that their use can enhance the routine decisions that people make every day, such as pricing products based on shifting demand and replenishing inventory as warehouse stocks are depleted.

> Agent uptake by industry has been slow but might be gaining momentum. ABS-SCM has been trialled in a wide range of sectors including high value manufacturing, retail and engineering services. Requirements that point towards an ABS solution need to be clear to prevent a solution that does not match expectations. The cost of deployment, once an important concern, is set to become lower with open source industrial frameworks becoming available. On the other hand, extant commercial solutions are still bespoke, and suffer from scalability and interoperability issues. Whilst AI and data analytics in SCM has gained attention and several firms took up data science as a core employee skill, ABS requires a wider skillset that is not commonly available. Typical SCM solutions providers such as SAP and Oracle are starting to offer ABS technology and may play a crucial role in its uptake. Trust into decentralised operations needs to be built. These issues may gain more attention with increasing interest in AI and decentralisation.

## 10. Software tools and platforms

A key factor corresponding to the implementation of agent-based systems is the use of an agent development framework. Over the past two decades, many tools and platforms across several programming languages for rapid development of agent-systems have been proposed. Due to the limited space and the focus of this article, we, in the section, present a short review on the representative tools and platforms for developing agent systems, reviews on this could be found in the work such as Bordini (2006) and Kravari and Bassiliades (2015) and a more recent comprehensive one by Pal et al. (2020).

We present agent development platforms in Table 3, which contains four main categories of tools: java-based frameworks (e.g., JADE (Bellifemine, Caire and Greenwood, 2007), JACK (Winikoff, 2005), Jason (Bordini, Hübner and Wooldridge, 2007), Repast (North, Collier and Vos, 2006), MASON (Luke et al., 2005)), Python-based frameworks (e.g., PADE (Melo et al., 2019), MESA (Masad and Kazil, 2015)); and new emergent framework (e.g., Fetch.ai AEA (Favorito et al., 2019; Minarsch et al., 2020)). One of frameworks worth further mention is the new emergent framework -- Fetch.AI AEA -- an autonomous economic agent framework built on the fetch blockchain and can seamlessly integrate with AI features and IoT sensors. The autonomous agents can find one another through the Open Economic Framework (OEF) (Sheikh and Simpson, 2019) and communicate with each other using built-in or user-defined protocols (FIPA or P2P). With the support of the IoT and AI, autonomous agents can interact with the environment and act on behalf of its owners. Another noteworthy development includes PTC ThingWorx which is primarily targeted towards industrial applications, and enables agent integration to the IoT for device modelling and business logic definition.

Table 3. List of representative agent development tools across many programming languages.

| Name | Programming Language | License | Description |
|------|---------------------|---------|-------------|
| JADE | Java | Open source | JADE is a software framework for the implementation of multi-agent |

| | | | systems through a middleware that complies with the FIPA specifications and through a set of graphical tools that supports debugging and deployment. More information can be found in https://jade.tilab.com/. |
|---|---|---|---|
| JACK | Java | Commercial license | JACK is a framework for agent system development which uses the BDII model and provides its own Java-based plan language. Its applications consist of a collection of autonomous agents that take input from the environment and communicate with other agents. |
| Jason | AgentSpeak | Open source | Jason is the first fully-fledged interpreter for an extended version of AgentSpeak, one of most influential abstract languages based on the BDI architecture. It provides a platform for the development of multi-agent systems with many customisable features. It also incorporates agent-oriented programming (AOP) language. Further details can access http://jason.sourceforge.net/wp/. |
| Repast | Java, C++ | Open source | Repast is a family of agent-based modeling and simulation platforms. It consists of two main components: Repast Simphony and Repast for High Performance Computing. More detail can be found in https://repast.github.io/. |
| MASON | Java | Open source | MASON is a fast discrete-event multi-agent simulation library that was designed to support large simulations and provide functionality for lightweight simulation needs. More information can be found in https://cs.gmu.edu/~eclab/projects/mason/. |
| PADE | Python | Open source | PADE is a framework for development, execution and management of multi-agent systems environments of distributed computation. It is fully written by Python and the Twisted [2] for implementing the communications between the network nodes. Documentation and tutorial can be found in https://pade.readthedocs.io/en/latest/. |
| MESA | Python | Open source | A modular framework for building, analysing and visualising agent-based models. It aims to be the Python-3 NetLogo, Repast and MASON. |
| OpenAI Gym | Python | Open source | Gym is a toolkit for developing and comparing reinforcement learning algorithms. It contains a collection of environments that can be used to test intelligent agents' ability to solve a problem. More details can be found in https://gym.openai.com/. |
| Fetch.ai AEA | Python | Open source | Fetch.ai AEA (Autonomous Economic Agent) is a framework to provide tools for creating autonomous economic agents that can act on an owner's behalf, with limited or no interference, and whose goal is to generate economic value to its owner. It incorporates blockchain and AI algorithms and supports the use of IoT sensors. Documentation and tutorials can be found in https://docs.fetch.ai/aea/. |

A number of Python and Java based Agent based software development tools exist. Whilst historically, Java – based tools have been popular, Python is gaining momentum in recent publications. An interesting recent development is Fetch.ai, which offers an "Agent search engine", and incorporates Blockchain and IoT, making it aligned with current topics of interest in SCM.

---

[2] https://twistedmatrix.com/trac/

# 11. Discussion and Conclusions

## *11.1 Motivators and Opportunities*

Several findings emerge from our review of agent-based system based automation in Supply Chain Management, which in turn inform future research directions.

Although ABS based automation is frequently used as an efficient way to solve a wide range of problems (Vallejo et al., 2010), in the field of planning, real-time control, robotics and many other industrial areas, but its take-up in the field of supply chains has been slow. However, a number of factors may motivate an increased interest in the field.

Supply chain automation research has thus far mainly focussed on two key areas: elementary iterative processes that surround dyadic relations, such as eCommerce negotiation between buyers and suppliers, and operational, holistic automation of supply chain processes, such as ad-hoc supply chain configuration and inventory coordination across multiple tiers. Increased complexity, and scale of supply chain structures are the main motivators behind these applications areas. These motivators are likely to stay, especially in complex engineering firms. However, other motivators may come into play in the near future. For example, in the manufacturing engineering community there is increased interest in mass customization with distributed manufacturing networks. Here networks of companies form ad hoc supply chains to cater to highly custom, low volume demand by utilising excess capacity. The sporadic nature of this proposition necessitates investment into digital infrastructure to deal with increased effort that will have to be spent in contracting, given relatively low returns.  Similar dynamics may be observed in collaborative logistic processes and resource sharing problems. Automated solutions in these areas may be key to success, as they avoid transaction costs. Another driver may include resilience. As supply chain interdependencies increase, so does the risk of disruptions. Solutions that can automatically intervene may prove to be valuable. Finally, increased consumer awareness and pressure in traceability and transparency in supply chains may yield the need for automated solutions that gather data on supply chains, which is a field that software agents can be put to use.

## *11.2 Challenges to be addressed*

To achieve increased adoption, several challenges need to be addressed. The field would benefit from a more structured approach to design and trialling of ABS solutions. Research into architectural design and communication and negotiation protocols have not fed into one another, with the majority of authors each creating their own bespoke approaches, with little or no comparison with previous proposals. This is partly due to a lack of commonly accepted performance metrics, and benchmark problems. Another reason is that each development, despite addressing common problems such as inventory coordination, is contextualised slightly differently, resulting in "case studies" rather than comparable innovations. Following an approach that is rooted in building on commonalities rather than generating wholly bespoke solutions to only slightly different problems may be more beneficial in speeding up development in this field.

Agent based system research in the field of Computer Science, despite having been motivated by applications such as eCommerce, has not been adopted by SCM researchers. Extensions of the Contract Net Protocol,  agent communication languages, and mobile agent technology are some examples that SCM may benefit from exploration.

Researchers have experimented with both collective and individual goal seeking behaviour in agents, however the contexts in which these propositions are false propositions and which are commonsensical are not clear. Although a wealth of supply chain data shows that collective behaviour is beneficial in mitigating unwanted emergent effects, collective goal seeking and data sharing usually are behavioural traits that are based in long term, strong relationships. If ABS is to be adopted by industry, systems that allow both collective and individual goal seeking behaviour need to be designed. If easy-to-configure systems are created, this itself may promote cooperation between members of the supply chain, and find new application fields such as logistics.

Related to this, communication and system integration needs to be paid particular attention. Researchers have developed wrappers for agent integration into ERP systems as a way to communicate between organisations. More research needs to be carried out to analyse the security, robustness and speed implications of such proposals as well as change management processes. As smart contracts (Szabo, 1997) and distributed ledger based transaction systems are gaining traction in the business world, integration of agents and these systems should be considered.

Developments in machine learning and predictive analytics have not yet found their way into ABS-SCM, however, learning and predictive agents have made much headway in the field of computer science, and separately, are gaining traction in SCM problems. We envisage these to play a role in SCM applications, particularly in negotiation strategy learning, inventory optimisation, learning to communicate and coordinate, inducing cooperative behaviour, and supply chain knowledge discovery.

Explainability will play a crucial role in the adaptation of these systems both on a micro scale where agent learning needs to be understood, and at a macro scale where collective action emerging from bottom-up needs to be traced.

Related to this, systems need to be highly reliable and have fallback strategies that ensure the failure of one agent will not affect other agents. There is no extant research on error handling ABS-SCM. Agent specific standards such as ISO/International Electrotechnical Commission (IEC), and Foundation for Intelligent Physical Agents (FIPA) and Mobile Agent Systems Interoperability Facility (MASIF) need to be investigated for their suitability into supply chain applications and emerging standards such as Reference Architectural Model for Industrie 4.0 (RAMI4.0) need to be analysed to determine how agents can communicate with other Industrie 4.0 components.

Finally, a number of common supply chain digitisation considerations apply also to agent-based automation. These include cost determination and responsibilities of setting up and maintaining common infrastructure with an understanding of associated return on investment.

Weiming Shen (2002) 'Distributed manufacturing scheduling using intelligent agents', IEEE intelligent systems, 17(1), pp. 88–94.

Winikoff, M. (2005) 'Jack<sup>TM</sup> Intelligent Agents: An Industrial Strength Platform', in Bordini, R. H. et al. (eds) Multi-Agent Programming: Languages, Platforms and Applications. Boston, MA: Springer US, pp. 175–193.

Wong, T. N. et al. (2006) 'Dynamic shopfloor scheduling in multi-agent manufacturing systems', Expert systems with applications, 31(3), pp. 486–494.

Wong, T. N. and Fang, F. (2010) 'A multi-agent protocol for multilateral negotiations in supply chain management', International Journal of Production Research, 48(1), pp. 271–299.

Wood, G. (2014) 'Ethereum: A secure decentralised generalised transaction ledger', Ethereum project yellow paper, 151(2014), pp. 1–32.

Wooldridge, M., Jennings, N. R. and Kinny, D. (2000) 'The Gaia methodology for agent-oriented analysis and design', Autonomous agents and multi-agent systems, 3(3), pp. 285–312.

Wooldridge, M. J. and Jennings, N. R. (1995) 'Intelligent agents: Theory and practice', Knowledge Engineering Review, 10(2), pp. 115–152.

Xiaocong Fan et al. (2006) 'MALLET - a multi-agent logic language for encoding teamwork', IEEE transactions on knowledge and data engineering, 18(1), pp. 123–138.

Xue, X. et al. (2005) 'An agent-based framework for supply chain coordination in construction', Automation in Construction, 14(3), pp. 413–430.

Ying, W. and Dayong, S. (2005) 'Multi-agent framework for third party logistics in E-commerce', Expert systems with applications, 29(2), pp. 431–436.

Yu, C. and Wong, T. N. (2015) 'A multi-agent architecture for multi-product supplier selection in consideration of the synergy between products', International Journal of Production Research, 53(20), pp. 6059–6082.

Yung, S. K. and Yang, C. C. (1999) 'A new approach to solve supply chain management problem by integrating multi-agent technology and constraint network', in Proceedings of the 32nd Annual Hawaii International Conference on Systems Sciences. 1999. HICSS-32. Abstracts and CD-ROM of Full Papers. Maui, HI, USA: IEEE, p. 10–pp.

Zarandi, M. H. F., Pourakbar, M. and Turksen, I. B. (2008) 'A Fuzzy agent-based model for reduction of bullwhip effect in supply chain systems', Expert systems with applications, 34(3), pp. 1680–1691.

Zhang, W. et al. (2015) 'An agent-based peer-to-peer architecture for semantic discovery of manufacturing services across virtual enterprises', Enterprise Information Systems, 9(3), pp. 233–256.